%% file: acl2023.tex
\pdfoutput=1

\documentclass[11pt]{article}

\usepackage[]{ACL2023}

\usepackage{times}
\usepackage{latexsym}
\usepackage{xspace}

\usepackage{color}
\usepackage{amsmath}
\usepackage{booktabs}
\usepackage{graphicx,lipsum} %
\usepackage[T1]{fontenc}

\usepackage[utf8]{inputenc}

\usepackage{microtype}

\usepackage{inconsolata}
\usepackage{multirow}

\newcommand{\dsname}{Bhasha-Abhijnaanam\xspace}

\newcommand{\modelname}{IndicLID\xspace}

\newcommand{\modelnamefasttextnative}{IndicLID-FTN\xspace}
\newcommand{\modelnamefasttextroman}{IndicLID-FTR\xspace}
\newcommand{\modelnameindicbert}{IndicLID-BERT\xspace}

\newcommand{\cbullet}{\noindent$\bullet$\;}

\title{\dsname: Native-script and Romanized Language Identification for 22 Indic Languages}

\author{Yash Madhani$^1$ \quad Mitesh M. Khapra$^2$ \quad
         Anoop Kunchukuttan$^3$ \\
        AI4Bharat$^{1,2,3}$ \quad IIT Madras$^{1,2,3}$ \quad Microsoft$^{3}$ \\
        {\tt $^{1}$cs20s002@cse.iitm.ac.in} \quad {\tt $^2$miteshk@cse.iitm.ac.in} \quad 
    {\tt $^{3}$ankunchu@microsoft.com} 
}

\begin{document}
\maketitle
\begin{abstract}


We create publicly available language identification (LID) datasets and models in all 22 Indian languages listed in the Indian constitution in both native-script and romanized text. First, we create \textit{\dsname}, a language identification test set for native-script as well as romanized text which spans all 22 Indic languages. We also train \textit{\modelname}, a language identifier for all the above-mentioned languages in both native and romanized script. For native-script text, it has better language coverage than existing LIDs and is competitive or better than other LIDs. \modelname is the first LID for romanized text in Indian languages. Two major challenges for romanized text LID are the lack of training data and low-LID performance when languages are similar. We provide simple and effective solutions to these problems. In general, there has been limited work on romanized text in any language, and our findings are relevant to other languages that need romanized language identification. Our models are publicly available at \url{https://ai4bharat.iitm.ac.in/indiclid} under open-source licenses. Our training and test sets are also publicly available at  \url{https://ai4bharat.iitm.ac.in/bhasha-abhijnaanam} under open-source licenses.

\end{abstract}

\section{Introduction}

In this work, we focus on building a language identifier for the 22 languages listed in the Indian constitution. With increasing digitization, there is a push to make NLP technologies like translation, ASR, conversational technologies, etc. \cite{bose2022bhashini} available as a public good at population scale \cite{chandorkar2022dpi}. A good language identifier is required to help build corpora in low-resource languages. For such languages, language identification is far from a solved problem due to noisy web crawls, small existing  datasets, and similarity to high-resource languages \cite{caswell-etal-2020-language}. 

Existing publicly available LID tools like CLD3\footnote{https://github.com/google/cld3}, LangID\footnote{https://github.com/saffsd/langid.py} \cite{lui-baldwin-2011-cross}, FastText\footnote{https://fasttext.cc/docs/en/language-identification.html} \cite{joulin2016fasttext} and NLLB\footnote{https://github.com/facebookresearch/fairseq/tree/nllb\#lid-model} \cite{nllb2022} have some shortcomings with respect to Indian languages. They do not cover all the above-mentioned 22 languages. In social media and chats, it is also common to use the roman script for most Indian languages leading to substantial user-generated content in roman script. However, none of the LIDs have any support for the detection of romanized Indian language text (except \texttt{cld3} support for Latin Hindi). The widespread use of romanization implies that accurate romanized Language Identification models are a critical component in the NLP stack for Indian languages, given that this affects over 735 million internet users \cite{KPMG-Report}. Therefore, our work on developing accurate and effective romanized Language Identification models has the potential to make a significant impact in the NLP space for Indian languages, particularly in the social media and chat application domains.
Hence, we undertake the task of creating a LID for these 22 Indian languages.  The main contributions of our work are as follows: 

\cbullet We create \textit{\dsname}\footnote{The word means language-identification in Sanskrit.}, a language identification test set for native-script as well as romanized text which spans 22 Indic languages. Previous benchmarks for native script do not cover all these languages \cite{nllb2022,roark2020processing}. The Dakshina test set for romanized text covers only 11 languages and there are ambiguous instances in the test set like named entities that cannot be assigned to a particular language \cite{roark2020processing}. 

\cbullet  We also train, \textit{\modelname}, an LID for all the above-mentioned languages in both native and romanized script. For native-script training data, we sample sentences from diverse sources and oversample low-resource languages. \modelname native-script model has better language coverage than existing LIDs and is competitive or better than other LIDs with 98\% accuracy and at least 6 times better throughput. 

\cbullet To the best of our knowledge, ours is one of the first large-scale efforts for romanized LID in any language, a task that has not received much attention. A major challenge for romanized text LID is the lack of romanized training data. We show that synthetic romanized training data created via transliteration can help train a reasonably good LID for romanized text. A simple linear classifier does not perform well for romanized text. Hence, we combine a simple but fast text classifier with a slower but more accurate classifier based on a pre-trained language model to achieve a good trade-off between accuracy and speed.  

Our findings are relevant to other languages that need LID for romanized text. We require native script data and a transliteration model to create the synthetic romanized data for the target language. This romanized data serves as training data for the romanized LID.




\input{Table_Bhasha_Abhijnaanam_stats}
\section{\dsname benchmark}

We describe the creation of the \dsname LID benchmark for 22 Indian languages in native and roman script. Table \ref{tab:Table_Bhasha_Abhijnaanam_stats} describes the statistics of the \textit{\dsname} benchmark. We build upon existing benchmarks to fill in the coverage and quality gaps and cost-efficiently cover all languages. 


\subsection{Native script test set.} We compile a native script test set comprising 19 Indian languages and 11 scripts from the FLORES-200 devtest \cite{nllb2022} and Dakshina sentence test set \cite{roark2020processing}. We create native text test sets for the remaining three languages (\textit{Bodo, Konkani, Dogri}) and one script (\textit{Manipuri} in \textit{Meetei Mayek} script) not covered in these datasets. For these new languages we first sample the English sentences from Wikipedia and ask in-house, professional translators to translate the sentences to respective languages. This method ensured the quality and accuracy of our test samples, as well as minimizing any potential noise in the data.\\
\subsection{Roman script test set.} We propose a new benchmark test set to evaluate roman-script language identification for 21 Indian languages. Out of these, 11 languages are represented in the Dakshina romanized sentence test set \cite{roark2020processing}, which comprises native script sentences from Wikipedia along with their romanization. However, this test set includes short sentences which are just named entities and English loan words which are not useful for romanized text LID evaluation.  
 To address this issue, we manually validate the Dakshina test sets for the languages we are interested in and filter out about 7\% of the sentences. Section \ref{sec:dakshina_filtering} describes the details of the filtering process. 
To create a benchmark test set for the remaining 10 Indian languages, we sampled sentences from IndicCorp \cite{doddapaneni2022indicxtreme} and asked annotators to write the same in roman script. We did not specify any transliteration guidelines and annotators were free to transliterate in the most natural way they deemed fit. We additionally asked annotators to skip the sentence if they find it invalid (wrong language, offensive, truncated, etc.).

\subsection{Romanized Dakshina testset filtering}
\label{sec:dakshina_filtering}

The Dakshina romanized sentence test set includes short sentences which are just named entities and English loan words which are not useful for romanized text LID evaluation. To address this issue, we manually validate the Dakshina test sets for the languages we are interested in. We first identified potentially problematic sentences from the romanized Dakshina test set by applying two constraints: (i) sentences shorter than 5 words, and (ii) native LID model is less confident  about the native language sentence (prediction score less than 0.8). These sentences were then validated by native language annotators. The annotators were asked to read the roman sentences and determine whether they were named entities or sentences where they could not determine the language. Such entries were filtered out. About 7\% of the sentences were filtered. Table \ref{tab:Table_Dakshina_filter_roman_stats_roman} describes the filtering statistics. 
\input{Table_Dakshina_filter_roman_stats_roman}

\section{\modelname Model}

\begin{figure}
\begin{center}
     \centering
     \includegraphics[scale = 0.3]{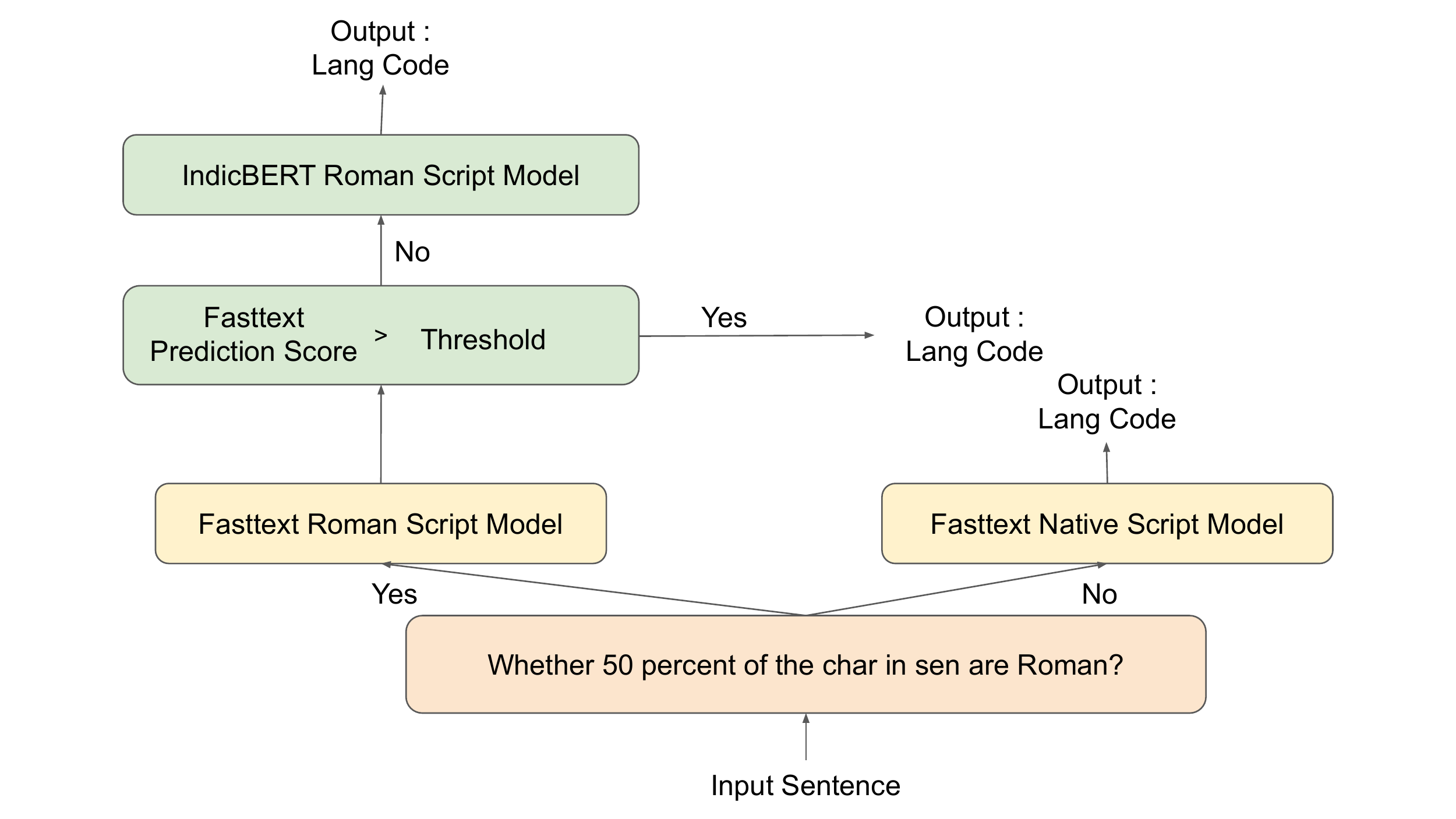}
    \caption{IndicLID Classifier Workflow}
    \label{fig:IndicLID_croped}
\end{center}
\end{figure}

IndicLID is a classifier specifically for Indic languages that can predict 47 classes (24 native-script classes and 21 roman-script classes plus English and Others). We create three classifier variants: a fast linear classifier, a slower classifier finetuned from a pre-trained LM, and an ensemble of the two models which trades off speed v/s accuracy. 

\subsection{Training dataset creation}
\noindent\textbf{Native-script training data.} We compiled the training data sentences from various sources viz. IndicCorp \cite{doddapaneni2022indicxtreme}, NLLB \cite{nllb2022}, Wikipedia,  Vikaspedia \footnote{https://vikaspedia.in} and internal sources. To ensure a diverse and representative training dataset, we sampled 100k sentences per language-script combination in a balanced way across all these sources. We used oversampling for languages with less than 100k sentences. We tokenized and normalized the sentences using IndicNLP library \footnote{https://github.com/anoopkunchukuttan/indic\_nlp\_library} \cite{kunchukuttan2020indicnlp} with default settings.



\noindent\textbf{Romanized training data.} 
There is hardly any romanized corpora for Indian languages in the public domain\footnote{CC-100 has romanized versions for 4 Indian languages, but a manual analysis suggested that it contains a lot of profane content.}. Hence, we explored the use of transliteration for creating synthetic romanized data. We create romanized training data by transliterating the native script training data into roman script using the multilingual IndicXlit\footnote{https://github.com/AI4Bharat/IndicXlit} transliteration model (Indic-to-En version) \cite{madhani2022aksharantar}, The authors have provided results on the transliteration quality of the IndicXlit model. We rely on this analysis to ensure the quality of generated training data. 


\subsection{Linear classifier}\label{sec:fasttext-model}
Linear classifiers using character n-gram features are widely used for LIDs \cite{jauhiainen2021comparing}. We use FastText \cite{joulin2016fasttext} to train our fast, linear classifier. It is a lightweight and efficient linear classifier that is well-suited for handling large-scale text data. It utilizes character n-gram features which enables it to utilize sub-word information. This makes it particularly useful for dealing with rare words and allows it to discriminate between similar languages having similar spellings. We trained separate classifiers for native script (\textbf{\modelnamefasttextnative}) and roman script (\textbf{\modelnamefasttextroman}). We chose 8-dimension word-vector models after experimentation as they maintain small model sizes without losing model accuracy (refer Appendix \ref{sec:hyperparameter_tuning_for_fastttext_roman} for results). 

\subsection{Pretrained LM-based classifier}\label{sec:language-model}
For romanized text, we observed that linear classifiers do not perform very well. Hence, we also experimented with models having larger capacity. Particularly, we finetuned a pretrained LM on the romanized training dataset. We evaluated the following LMs: XLM-R \cite{conneau-etal-2020-unsupervised}, IndicBERT-v2 \cite{doddapaneni2022indicxtreme} and MuRIL \cite{khanuja2021muril}. The last two LMs are specifically trained for Indian languages and MuRIL also incorporates synthetic romanized data in pre-training. Hyperparameters for finetuning are described in Appendix \ref{sec:model_selection_LM}.  We used IndicBERT-based classifier as the LM-based classifier (henceforth referred to as \textbf{\modelnameindicbert}) since it was amongst the best-performing romanized text classifiers and had maximum language coverage.


\subsection{Final Ensemble classifier}

Our final \modelname classifier is an pipeline of multiple classifiers. Figure \ref{fig:IndicLID_croped} shows the overall workflow of the \modelname classifier. The pipeline works as described here: (1) Depending on the amount of roman script in the input text, we invoke either the native-text or romanized linear classifier. \modelnamefasttextroman  is invoked for text containing $>$50\% roman characters. (2) For roman text, if \modelnamefasttextroman is not confident about its prediction, we redirect the request to the \modelnameindicbert. We resort to this two-stage approach for romanized input to achieve a good trade-off between classifier accuracy and inference speed. The fast \modelnamefasttextroman's prediction is used if the model is confident about its prediction (probability of predicted class $>0.6$ ), else the slower but  more accurate \modelnameindicbert is invoked. This threshold provides a good trade-off (See Appendix \ref{sec:analysis-speed/accuracy-tradeoff} for more details).

\section{Results and Discussion}
We discuss the performance of various models on the benchmark and analyze the results. To prevent any overlap between the test/valid and train sets, we excluded the Flores-200 test set \cite{nllb2022}, Dakshina test set \cite{roark2020processing} while sampling native train samples from various sources. Additionally, we removed the training samples from the benchmark samples when collecting sentences for the benchmark test set. We also made sure that there was no overlap between the test and valid sets.
To create the romanized training set, we simply transliterated the native training set. As the Dakshina test set \cite{roark2020processing} provided parallel sentences for the native and roman test sets, there was no overlap between the roman train and test sets.

\input{Table_IndicLID_performance_comparison_nllb_cld3_native_script}

\subsection{Native script LID}
We compare \modelnamefasttextnative with the NLLB model \cite{nllb2022} and the CLD3 model. As we can see in Table \ref{tab:Table_IndicLID_performance_comparison_nllb_cld3_native_script}, the LID performance of \modelnamefasttextnative is comparable or better than other models. Our model is 10 times faster and 4 times smaller than the NLLB model. The model's footprint can be further reduced by model quantization \cite{joulin2016fasttext} which we leave for future work.    


\subsection{Roman script LID}\label{sec:roman-model-result-and-analysis}
Table \ref{tab:Table_IndicLID_performance_roman_script} presents the results of different model variants on the romanized test set (see Appendix \ref{sec:language-wise-analysis} for language-wise results).  \modelnameindicbert is significantly better than \modelnamefasttextroman, but the throughput decreases significantly. The ensemble model (\modelname) maintains the same LID performance as \modelnameindicbert with a 3x increase in the throughput over \modelnameindicbert. Further speedups in the model throughput can be achieved by creating distilled versions, which we leave for future work. 

\begin{figure}
\begin{center}
     \centering
     \includegraphics[scale = 0.20]{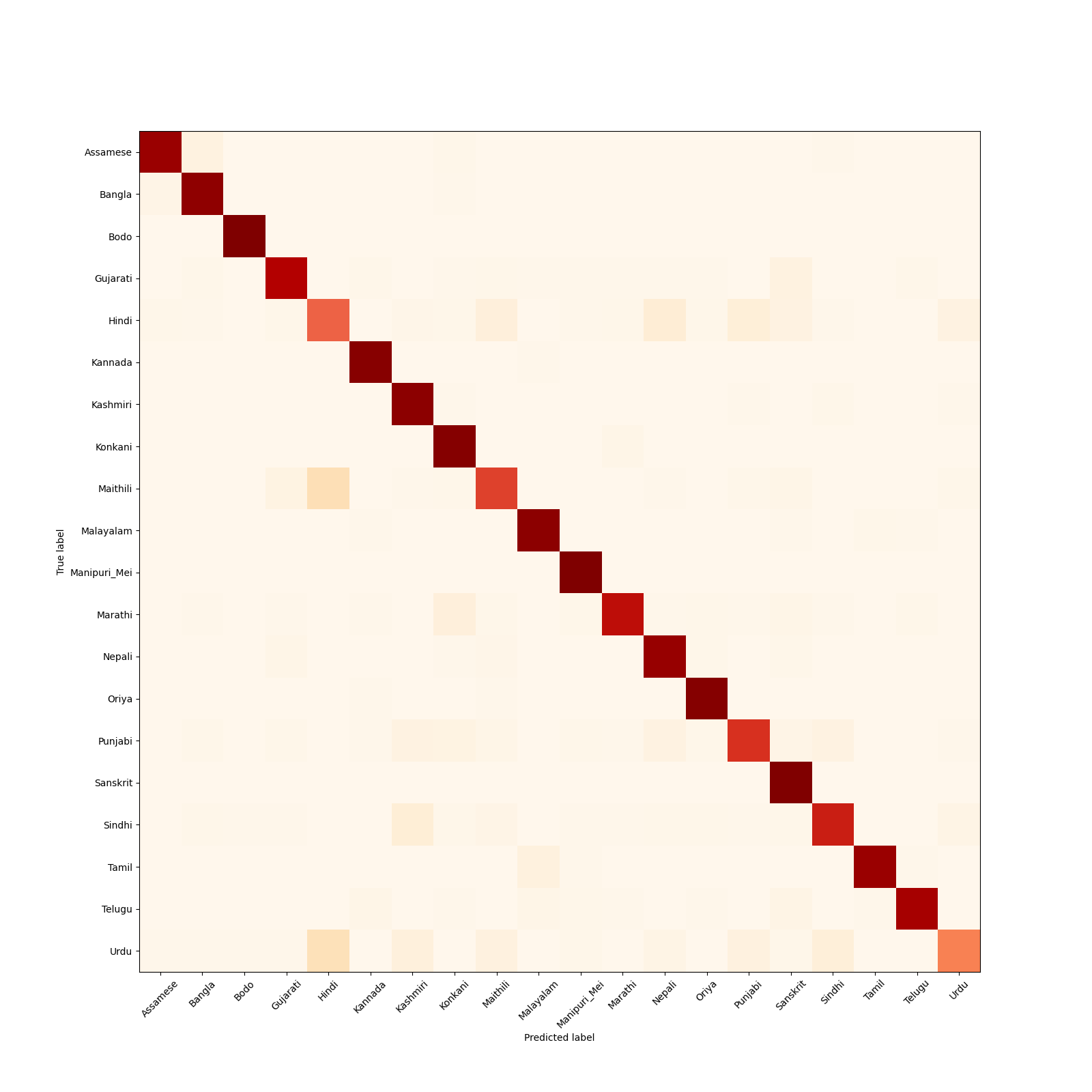}
    \caption{Confusion matrix (\modelname, roman testset)}
    \label{fig:confusion_matrix_test_combine_roman_without_values_20_langs}
\end{center}
\end{figure}
\input{Table_IndicLID_performance_roman_script}


\noindent\textbf{LID confusion analysis} The confusion matrix for \modelname is shown in Figure \ref{fig:confusion_matrix_test_combine_roman_without_values_20_langs}. We see that major confusions are between similar languages. Some examples of such language clusters that can be observed are (1) Hindi and very close languages like Maithili, Urdu and Punjabi, (2) Konkani and Marathi, (3) Sindi and Kashmiri. Improving romanized LID between very similar languages is thus an important direction of improvement.

\input{Table_synthetic_data_impact_roman_vs_romanized_benchmark}

\noindent\textbf{Impact of synthetic training data} To understand the impact of synthetic training data, we generate a machine-transliterated version of the romanized test set using IndicXlit. We compare the LID accuracy on the original and synthetically generated test sets. Table \ref{tab:Table_synthetic_data_impact_roman_vs_romanized_benchmark} shows that the results on the synthetic test set are significantly better than the original test set (approaching accuracy levels in the 90s). The data characteristics of the synthetic test set are much closer to the training data than the original test set. Closing the training-test distribution gap (by representing original romanized data in the training data and/or improved generation of synthetic romanized data to reflect true data distribution) is critical to improving model performance.

The confusion matrix gives further insights into the impact of synthetic training data. Hindi is confused with languages like  Nepali, Sanskrit, Marathi and Konkani using the same native script as Hindi (Devanagari). Since a multilingual transliteration model with significant Hindi data was used to create the synthetic romanized training data, it  may result in the synthetic romanized forms of these languages being more similar to Hindi than would be the case with original romanized data. 

\begin{figure}
\begin{center}
     \centering
     \includegraphics[width=0.45\textwidth,height=0.20\textwidth]{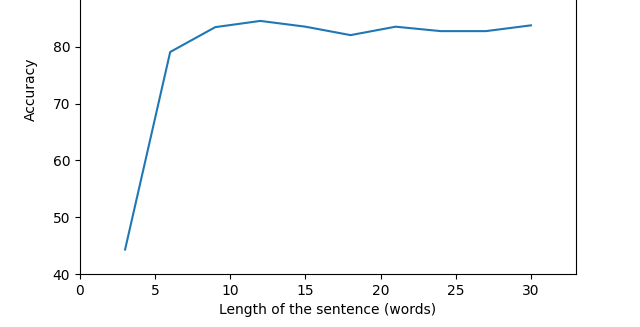}
    \caption{Effect of input length on romanized testset}
    \label{fig:len_wise_accuracy}
\end{center}
\end{figure}

\noindent\textbf{Impact of input length} Figure \ref{fig:len_wise_accuracy} plots the LID accuracy for various input length buckets. The LID is most confused for short inputs (<10 words) after which the performance is relatively stable.

\section{Conclusion}
We introduce an LID benchmark and models for native-script and romanized text in 22 Indian languages. These tools will serve as a basis for building NLP resources for Indian languages, particularly extremely low-resource ones that are "left-behind" in the NLP world today \cite{joshi-etal-2020-state}. Our work takes first steps towards LID of romanized text, and our analysis reveals directions for future work.  

\section*{Acknowledgements}
We would like to thank the Ministry of Electronics and Information Technology of the Government of India for their generous grant through the Digital India Bhashini project. We also thank the Centre for Development of Advanced Computing for providing compute time on the Param Siddhi Supercomputer. We also thank Nilekani Philanthropies for their generous grant towards building datasets, models, tools and resources for Indic languages. We also thank Microsoft for their grant to support research on Indic languages. We would like to thank Jay Gala and Ishvinder Sethi for their help in coordinating the annotation work. Most importantly we would like to thank all the annotators who helped create the \dsname benchmark.

\section*{Limitations}
The benchmark for language identification for the most part contains clean sentences (grammatically correct, single script, etc.). Data from the real world might be noisy  (ungrammatical, mixed scripts, code-mixed, invalid characters, etc.). A better representative benchmark might be useful for such use cases. However, the use cases captured by this benchmark should suffice for the collection of clean monolingual corpora. This also represents a first step for many languages where no LID benchmark exists. 

The use of synthetic training data seems to create a gap in performance due to divergence in train/test data distributions. Acquisition of original native romanized text and methods to generate better romanized text are needed.  

Note that the romanized LID model does not support Dogri since the IndicXlit transliteration model does not support Dogri. However, since Dogri is written in the Devanagari script using the transliterator for Hindi which uses the same script might be a good approximation to generate synthetic training data. We will explore this in the future.    

This work is limited to the 22 languages listed in the 8$^{th}$ schedule of the Indian constitution. Further work is needed to extend the benchmark to many more widely used languages in India (which has about 30 languages with more than a million speakers).



\section*{Ethics Statement}
For the human annotations on the dataset, the language experts are native speakers of the languages and from the Indian subcontinent. They were paid a competitive monthly salary to help with the task. The salary was determined based on the skill set and experience of the expert and adhered to the norms of the government of our country. The dataset has no harmful content. The annotators were made aware of the fact that the annotations would be released publicly and the annotations contain no private information. The proposed benchmark builds upon existing datasets. These datasets and related works have been cited. 

The annotations are collected on a publicly available dataset and will be released publicly for future use. The IndicCorp dataset which we annotated has already been checked for offensive content. 

All the datasets created as part of this work will be released under a CC-0 license\footnote{\url{https://creativecommons.org/publicdomain/zero/1.0}} and all the code and models will be released under an MIT license.\footnote{\url{https://opensource.org/licenses/MIT}}




\bibliography{anthology,custom}
\bibliographystyle{acl_natbib}

\appendix




\section{Hyperparameter tuning for Roman script linear classifier }
\label{sec:hyperparameter_tuning_for_fastttext_roman} 
We train the \modelnamefasttextroman model using 100k samples. While deciding the configuration \modelnamefasttextroman model, we experimented with fixing the dimension of \modelnamefasttextroman model and tuning on the rest of the hyperparameters. As we can see from table \ref{tab:Table_Fasttext_with_different_dimension_clean_roman} model size increases with the increase of \modelnamefasttextroman dimension. However, beyond 8 dimensions, there is not much improvement observed. Therefore, we chose the model with 8 dimensions, taking into account the model size. 
\input{Table_Fasttext_with_different_dimension_clean_roman}
\input{Table_finetuning_different_LM_with_freezing_all_layers_roman}

\input{Table_IndicBERT_finetuning_unfreezing_different_layers_roman}

\section{Model selection for Roman script LM-based classifier}
\label{sec:model_selection_LM} 
We experimented with three different pre-trained language models: IndicBERT \cite{doddapaneni2022indicxtreme}, XLM-R \cite{conneau-etal-2020-unsupervised}, and MuRIL \cite{khanuja2021muril}. In the initial experiment, we froze all the layers except for the last softmax layer and finetuned the model with our training data. To fine-tune the language model, we added one softmax layer to the end of the model and used our roman script training data to finetune the model. The results for these experiments are shown in Table \ref{tab:Table_finetuning_different_LM_with_freezing_all_layers_roman}. We found that IndicBERT and MuRIL performed similarly among these three models for our roman LID task. MuRIL leverages the advantage of roman text training data, while IndicBERT was trained on the only native script but performed similarly. However, IndicBERT supports 24 Indian languages, while MuRIL only supports 17 Indian languages. Therefore, we selected IndicBERT due to its superior coverage and performance. \\
We then further experimented with IndicBERT by unfreezing 1, 2, 4, 6, 8, and 11 layers. The results and comparison of all the experiments are described in Table \ref{tab:Table_IndicBERT_finetuning_unfreezing_different_layers_roman}. We found that unfreezing 1 layer was enough for our task and that unfreezing more layers did not provide any additional benefit.

\input{Table_comparing_different_threshold_speed_vs_accuracy_roman}

\section{Analysis of speed/accuracy tradeoff}
\label{sec:analysis-speed/accuracy-tradeoff}
We experimented \modelname with different thresholds. If the probability score is below a certain threshold we invoke a more powerful model \modelnameindicbert, otherwise, we go with \modelnamefasttextroman model prediction. \modelnamefasttextroman model is quite fast as compared to \modelnameindicbert model. We can see a good trade-off between throughput and accuracy in table \ref{tab:Table_comparing_different_threshold_speed_vs_accuracy_roman} as we increase the threshold. As the threshold increases, the input is more likely to go towards the \modelnameindicbert model, as we are making the model less reliant on the \modelnamefasttextroman model.

\section{Language-wise analysis for Roman script classifiers}
\label{sec:language-wise-analysis} 
Table \ref{tab:Table_Fasttext_IndicBERT_IndicLID_f1_precision_recall_roman} illustrates the language-specific performance of \modelnamefasttextroman, \modelnameindicbert and \modelname models in detail. As we can see \modelnameindicbert has better representation than \modelnamefasttextroman for almost all the languages which leads better F1 score for \modelname. However, for the languages of Sanskrit and Manipuri, the \modelnamefasttextroman model has a better representation than the \modelnameindicbert model, which is an interesting finding that warrants further investigation in future studies.
\input{Table_Fasttext_IndicBERT_IndicLID_f1_precision_recall_roman}

\end{document}

%% file: Table_Bhasha_Abhijnaanam_stats.tex
\begin{table}[t]
    \centering
    \scalebox{0.84}{
    \begin{tabular}{llrr}
     \toprule
    \textbf{Language} &\textbf{Script} &\textbf{Native} &\textbf{Roman} \\
    \midrule
    Assamese &Bengali &1012 &\textbf{512} \\
    Bangla &Bengali &5606 &4595 \\
    Bodo &Devanagari &\textbf{1500} &\textbf{433} \\
    Dogri &Devanagari &\textbf{1498} &\textbf{512} \\
    Gujarati &Gujarati &5797 &4785 \\
    Hindi &Devanagari &5617 &4606 \\
    Kannada &Kannada &5859 &4848 \\
    \multirow{2}{*}{Kashmiri} &Perso-Arabic &2511 & \multirow{2}{*}{\textbf{450}} \\
     &Devanagari &1012 & \\
    Konkani &Devanagari &\textbf{1500} &\textbf{444} \\
    Maithili &Devanagari &2512 &\textbf{439} \\
    Malayalam &Malayalam &5628 &4617 \\
    \multirow{2}{*}{Manipuri} &Bengali &1012 &  \multirow{2}{*}{\textbf{442}}\\
     & Meetei Mayek &\textbf{1500} & \\ 
    Marathi &Devanagari &5611 &4603 \\
    Nepali &Devanagari &2512 &\textbf{423} \\
    Oriya &Oriya &1012 &\textbf{512} \\
    Punjabi &Gurmukhi &5776 &4765 \\
    Sanskrit &Devanagari &2510 &\textbf{448} \\
    Santali &Ol Chiki &2512 &0 \\
    Sindhi &Perso-Arabic &5893 &4881 \\
    Tamil &Tamil &5779 &4767 \\
    Telugu &Telugu &5751 &4741 \\
    Urdu &Perso-Arabic &6883 &4371 \\
    \bottomrule
    \end{tabular}
    }
    \caption{Summary of the \dsname benchmark. Number of romanized and native-script sentences are reported. The cells in \textbf{bold} indicate the datasets newly contributed by this work. Romanized Santali testset has not been created since Santhali annotators we contacted did not use roman script and spoke Bengali as a second language. \citet{nllb2022} also cite a similar experience.}
    \label{tab:Table_Bhasha_Abhijnaanam_stats}
\end{table}

%% file: Table_Dakshina_filter_roman_stats_roman.tex
\begin{table}
    \centering
    \scalebox{0.68}{
    \begin{tabular}{l|rrrrrr}
    \toprule
\textbf{Language} &\textbf{Total samples} &\textbf{Valid samples} &\textbf{\%filtered} \\
    \midrule
    Bengali &5001 &4600 &8.0183 \\
    Gujarati &5001 &4789 &4.2391 \\
    Hindi &5001 &4616 &7.6984 \\
    Kannada &5001 &4849 &3.0393 \\
    Malayalam &5001 &4627 &7.4785 \\
    Marathi &5001 &4617 &7.6784 \\
    Punjabi &5001 &4782 &4.3791 \\
    Sindhi &5001 &4889 &2.2395 \\
    Tamil &5001 &4802 &3.9792 \\
    Telugu &5001 &4754 &4.9390 \\
    Urdu &4881 &4395 &9.9569 \\
    \bottomrule
    \end{tabular}
    }
    \caption{Statistics of Dakshina roman filtered test set}
    \label{tab:Table_Dakshina_filter_roman_stats_roman}
\end{table}

%% file: Table_IndicLID_performance_comparison_nllb_cld3_native_script.tex
\begin{table}
\renewcommand{\arraystretch}{1.3}
    \centering
    \scalebox{0.56}{
    \begin{tabular}{lrrrrrr}
    \toprule
    \textbf{Model} &\textbf{P} &\textbf{R} &\textbf{F1} &\textbf{Acc} &\textbf{Throughput} &\textbf{Size} \\  
    \midrule
    \modelnamefasttextnative-8-dim (24) &98.11 &98.56 &98.31 &98.55 &30,303 &318M \\
    \midrule
    \multicolumn{7}{l}{\textit{Comparing our \modelnamefasttextnative model with CLD3 model (12)}} \\
    \modelnamefasttextnative-4-dim  &99.43 &98.40 &98.89 &98.33 &47,619 &208M \\
    \modelnamefasttextnative-8-dim  &99.73 &98.67 &99.18 &98.62 &33,333 &318M \\
    CLD3 &98.52 &98.14 &98.31 &98.03 &4,861 &-  \\
    \midrule
    \multicolumn{7}{l}{\textit{Comparing our \modelnamefasttextnative model with NLLB model (20)}} \\
    \modelnamefasttextnative-4-dim &97.78 &98.10 &97.92 &98.19 &41,666 &208M \\
    \modelnamefasttextnative-8-dim &98.13 &98.59 &98.34 &98.56 &29,411 &318M \\
    NLLB &99.28 &98.65 &98.95 &98.78 &4,970 &1.1G \\
    \bottomrule
    \end{tabular}
    }
    \caption{Benchmarking on the \dsname native-script testset. For fair comparison with NLLB and CLD3, we restrict the comparison to languages that are common with \modelnamefasttextnative (count of common languages is indicated in brackets). Throughput is number of sentence/second.}
    \label{tab:Table_IndicLID_performance_comparison_nllb_cld3_native_script}
\end{table}

%% file: Table_IndicLID_performance_roman_script.tex
\begin{table}
    \centering
    \scalebox{0.50}{
    \begin{tabular}{l|rrrrrr}
    \toprule
    \textbf{Model} &\textbf{P} &\textbf{R} &\textbf{F1} &\textbf{Acc} &\textbf{Throughput} &\textbf{Size} \\  
    \midrule
    \modelnamefasttextroman (dim-8) &63.12 &78.01 &63.28 &71.49 &37,037 &357 M \\
    \modelnameindicbert (unfeeze-layer-1) &72.70 &84.01 &74.52 &80.04 &3 &1.1 GB \\
    \modelname (threshold-0.6) &72.74 &84.50 &74.72 &80.40 &10 &1.4 GB \\
    \bottomrule
    \end{tabular}
    }
    \caption{Performance of \modelnamefasttextroman on \dsname roman script test set. Throughput is number of sentence/second.}
    \label{tab:Table_IndicLID_performance_roman_script}
\end{table}

%% file: Table_synthetic_data_impact_roman_vs_romanized_benchmark.tex
\begin{table}
    \centering
    \scalebox{0.72}{
    \begin{tabular}{l|rrrr}
    \toprule
    \textbf{Testset} &\textbf{P} &\textbf{R} &\textbf{F1} &\textbf{Acc} \\  
    \midrule
    Original &72.74 &84.50 &74.72 &80.40 \\
    Synthetic &90.79 &97.24 &93.43 &95.96 \\
    \bottomrule
    \end{tabular}
    }
    \caption{Comparison of results on Synthetic vs. original Romanized test sets for \modelname model}
    \label{tab:Table_synthetic_data_impact_roman_vs_romanized_benchmark}
\end{table}

%% file: Table_Fasttext_with_different_dimension_clean_roman.tex
\begin{table}[t]
    \centering
    \scalebox{0.50}{
    \begin{tabular}{l|rrrrrr}
    \toprule
    \textbf{Dimension} &\textbf{Precision} &\textbf{Recall} &\textbf{F1-Score} &\textbf{Accuracy} &\textbf{Throughput} &\textbf{Model Size} \\    
    \midrule
    4	&60.01	&74.56	&61.09	&67.52	&50000	&171M \\
    8	&63.13	&78.02	&63.29	&71.49	&37037	&357M \\
    16	&63.67	&78.33	&64.32	&71.58	&30303	&578M \\
    32	&64.62	&78.67	&65.16	&71.95	&15625	&1.6G \\
    64	&64.54	&78.58	&65.10	&71.93	&14085	&1.9G \\
    128	&64.55	&78.45	&65.03	&71.77	&9901	&3.3G \\
    256	&64.60	&78.54	&65.13	&71.89	&7463	&7.3G \\
    512	&63.89	&78.29	&64.58	&71.49	&4608	&11G \\
    768	&64.37	&78.63	&65.07	&72.04	&3876	&22G \\
    1024	&64.30	&78.53	&65.07	&71.94	&3322	&29G \\
    \bottomrule
    \end{tabular}
    }
    \caption{\modelnamefasttextroman performance on \dsname roman script test set. \modelnamefasttextroman are hyper-tuned by fixing different dimensions. Throughput is number of sentence/second.}
    \label{tab:Table_Fasttext_with_different_dimension_clean_roman}
\end{table}

%% file: Table_finetuning_different_LM_with_freezing_all_layers_roman.tex
\begin{table}
    \centering
    \scalebox{0.55}{
    \begin{tabular}{l|rrrr}
    \toprule
    \textbf{Model} &\textbf{Precision} &\textbf{Recall} &\textbf{F1-Score} &\textbf{Accuracy} \\  
    \midrule
    XLMR \cite{conneau-etal-2020-unsupervised} &63.19 &70.92 &59.49 &65.15 \\
    MuRIL \cite{khanuja2021muril} &66.70 &79.08 &67.77 &73.70 \\
    IndicBERT \cite{doddapaneni2022indicxtreme} &68.07 &80.52 &68.91 &75.81 \\
    \bottomrule
    \end{tabular}
    }
    \caption{\dsname roman script test set results on roman script Language models finetuned by freezing all the  layers}
    \label{tab:Table_finetuning_different_LM_with_freezing_all_layers_roman}
\end{table}

%% file: Table_IndicBERT_finetuning_unfreezing_different_layers_roman.tex
\begin{table}
    \centering
    \scalebox{0.70}{
    \begin{tabular}{l|rrrr}
    \toprule
    \textbf{Model} &\textbf{Precision} &\textbf{Recall} &\textbf{F1-Score} &\textbf{Accuracy} \\  
    \midrule
    unfreezed-layer-1	&72.70	&84.01	&74.53	&80.04 \\
    unfreezed-layer-2	&69.84	&83.84	&72.44	&79.55 \\
    unfreezed-layer-4	&69.53	&83.44	&72.12	&79.47 \\
    unfreezed-layer-6	&68.41	&81.89	&70.02	&77.08 \\
    unfreezed-layer-8	&67.46	&81.88	&68.42	&76.04 \\
    unfreezed-layer-11	&70.55	&83.73	&72.63	&79.88 \\
    \bottomrule
    \end{tabular}
    }
    \caption{\dsname roman script test set results on \modelnameindicbert finetuned with unfreezing different numbers of layers}
    \label{tab:Table_IndicBERT_finetuning_unfreezing_different_layers_roman}
\end{table}

%% file: Table_comparing_different_threshold_speed_vs_accuracy_roman.tex
\begin{table}
    \centering
    \scalebox{0.70}{
    \begin{tabular}{l|rrrrr}
    \toprule
    \textbf{Thresholds} &\textbf{P} &\textbf{R} &\textbf{F1} &\textbf{Acc} &\textbf{Throughput}\\
    \midrule
    threshold 0.1	&63.13	&78.02	&63.29	&71.49	&50000 \\
    threshold 0.2	&63.43	&78.18	&63.63	&71.77	&379 \\
    threshold 0.3	&65.50	&79.64	&66.15	&73.84	&54 \\
    threshold 0.4	&68.39	&81.84	&69.77	&76.84	&22 \\
    threshold 0.5	&70.99	&83.60	&72.87	&79.15	&14 \\
    threshold 0.6	&72.74	&84.51	&74.72	&80.4	&10 \\
    threshold 0.7	&73.60	&84.80	&75.54	&80.93	&9 \\
    threshold 0.8	&73.88	&84.81	&75.77	&80.96	&8 \\
    threshold 0.9	&73.51	&84.50	&75.35	&80.62	&6 \\
    \bottomrule
    \end{tabular}
    }
    \caption{Trade-off between inference time and accuracy with different thresholds. Throughput is number of sentence/second.}
    \label{tab:Table_comparing_different_threshold_speed_vs_accuracy_roman}
\end{table}

%% file: Table_Fasttext_IndicBERT_IndicLID_f1_precision_recall_roman.tex
\begin{table*}[t]
    \centering
    \scalebox{0.80}{
    \begin{tabular}{l|rrr|rrr|rrr}
    \toprule
    &\multicolumn{3}{c|}{\modelnamefasttextroman (8 dim)} &\multicolumn{3}{c|}{\modelnameindicbert (unfreeze 1)} &\multicolumn{3}{c}{\modelname (threshold 0.6)} \\
     \midrule
     &Precision &Recall &F1 &Precision &Recall &F1 &Precision &Recall &F1 \\
     \midrule
    Assamese	&37.72	&93.55	&53.76	&66.81	&91.21	&77.13	&72.41	&92.77	&\textbf{81.34} \\
    Bangla	&76.63	&94.10	&84.47	&97.12	&88.14	&92.41	&94.94	&93.95	&\textbf{94.44} \\
    Bodo	&70.88	&98.38	&82.40	&84.78	&99.08	&91.37	&85.66	&99.31	&\textbf{91.98} \\
    Konkani	&24.62	&95.72	&39.17	&38.35	&99.32	&55.33	&40.90	&97.75	&\textbf{57.67} \\
    Gujarati	&89.52	&78.70	&83.76	&95.88	&85.20	&90.23	&95.16	&86.69	&\textbf{90.73} \\
    Hindi	&65.46	&15.68	&25.29	&76.32	&60.40	&67.43	&77.16	&53.32	&\textbf{63.06} \\
    Kannada	&89.66	&96.41	&92.91	&95.79	&95.71	&95.75	&95.29	&96.78	&\textbf{96.03} \\
    Kashmiri	&18.74	&91.56	&31.12	&39.45	&93.11	&55.42	&34.80	&94.67	&\textbf{50.90} \\
    Maithili	&07.81	&38.95	&13.01	&29.00	&41.69	&34.21	&21.97	&43.74	&\textbf{29.25} \\
    Malayalam	&89.75	&94.46	&92.04	&92.19	&95.32	&93.73	&91.33	&95.36	&\textbf{93.30} \\
    Manipuri	&64.84	&98.87	&\textbf{78.32}	&50.06	&98.42	&66.36	&58.85	&99.32	&73.91 \\
    Marathi	&87.21	&79.58	&83.22	&96.35	&80.80	&87.89	&95.86	&82.92	&\textbf{88.92} \\ 
    Nepali	&19.55	&82.51	&31.61	&43.25	&93.85	&59.21	&36.94	&93.62	&\textbf{52.98} \\
    Oriya	&41.88	&95.70	&58.26	&64.09	&95.51	&76.71	&62.96	&97.27	&\textbf{76.44} \\
    Punjabi	&78.52	&37.21	&50.49	&84.71	&64.64	&73.32	&85.62	&62.62	&\textbf{72.34} \\
    Sanskrit	&49.32	&96.43	&\textbf{65.26}	&32.55	&99.33	&49.04	&36.88	&99.11	&53.75 \\
    Sindhi	&80.00	&61.05	&69.25	&86.39	&71.91	&78.49	&87.88	&72.51	&\textbf{79.46} \\
    Tamil	&97.32	&90.56	&93.82	&97.15	&93.06	&95.06	&97.50	&92.64	&\textbf{95.01} \\
    Telugu	&94.24	&87.68	&90.84	&95.25	&88.76	&91.89	&95.89	&89.50	&\textbf{92.58} \\
    Urdu	&78.88	&33.24	&46.77	&88.53	&44.84	&59.53	&86.87	&46.31	&\textbf{60.41} \\
    Avg	&63.13	&78.02	&63.29	&72.70	&84.01	&74.53	&72.74	&84.51	&\textbf{74.72} \\
    \bottomrule
    \end{tabular}
    }
    \caption{Precision, recall and F1-score of \modelnamefasttextroman, \modelnameindicbert and \modelname roman script model. All scores are calculated on \dsname roman script test set. \textbf{Bold} indicates the best language representation among \modelnamefasttextroman, \modelnameindicbert and \modelname roman script model for individual languages. }
    \label{tab:Table_Fasttext_IndicBERT_IndicLID_f1_precision_recall_roman}
\end{table*}